\documentclass[runningheads]{llncs}

\usepackage{graphicx}
\usepackage{hyperref}

\usepackage{amsmath} 
\usepackage{amssymb}
\usepackage{multirow}
\usepackage{doi}
\usepackage{makecell}
\usepackage{subcaption}
\usepackage{graphicx}
\usepackage{epstopdf}
\usepackage{here}
\usepackage{bbding}
\usepackage{booktabs}
\usepackage{multirow}
\usepackage[table,xcdraw]{xcolor}

\input alphabet
\def\Diag{\mathrm{Diag}}

\def\R{\mathbb{R}}

\usepackage{xcolor}
\definecolor{cadmiumgreen}{rgb}{0.0, 0.42, 0.24}
\newcommand{\YZ}[1]{\textcolor{black}{#1}}
\newcommand{\CH}[1]{\textcolor{black}{#1}}
\newcommand{\DM}[1]{\textcolor{black}{#1}}

\newcommand{\yz}[1]{\textcolor{black}{#1}}
\newcommand{\ch}[1]{\textcolor{black}{#1}}
\newcommand{\dm}[1]{\textcolor{black}{#1}}
\newcommand{\ji}[1]{\textcolor{black}{#1}}

\usepackage{soul}

\usepackage{comment}

\begin{document}

\title{Ultrasound Image Reconstruction with Denoising Diffusion Restoration Models}
\titlerunning{US Image Reconstruction with DDRM}

 \author{Yuxin Zhang \and Clément Huneau \and Jérôme Idier \and Diana Mateus} 
 \authorrunning{Y. Zhang et al.}
 \institute{Nantes Université, École Centrale Nantes, LS2N,\\ CNRS, UMR 6004, F-44000 Nantes, France \\ \email{yuxin.zhang@ls2n.fr}}


\maketitle            

\begin{abstract}
Ultrasound image reconstruction can be approximately cast as a linear inverse problem that has traditionally been solved with penalized optimization using the $l_1$ or $l_2$ norm, or wavelet-based terms. However, such regularization functions often struggle to balance the sparsity and the smoothness. A promising alternative is using learned priors to make the prior knowledge closer to reality. 
In this paper, we rely on learned priors under the framework of Denoising Diffusion Restoration Models (DDRM), initially conceived for restoration tasks with natural images. 
We propose and test two adaptions of DDRM to ultrasound inverse problem models, DRUS and WDRUS. Our experiments on synthetic and PICMUS data show that from a single plane wave our method can achieve image quality comparable to or better than DAS and state-of-the-art methods. 
The code is available at: \href{https://github.com/Yuxin-Zhang-Jasmine/DRUS-v1}{https://github.com/Yuxin-Zhang-Jasmine/DRUS-v1}.

\end{abstract}

\section{Introduction}
Ultrasound (US) imaging is a popular non-invasive imaging modality, of
wide\-spread use in medical diagnostics due to its safety and cost-effectiveness tradeoff. Standard commercial scanners rely on simple beamforming algorithms,  e.g.\ Delay-and-Sum (DAS), to transform raw signals into B-mode images, trading spatial resolution for speed. Yet, many applications could benefit from improved resolution and contrast, enabling better organ and lesion boundary detection.

Recent techniques to improve US image quality include adaptive beamforming techniques, e.g.\ based on Minimum Variance (MV) estimation \cite{synnevag_adaptive_2007,asl_eigenspace-based_2010}, or Fourier-based reconstructions~\cite{Chernyakova-Eldar_2018}. Other methods focus on optimizing either pre-\cite{REFOCUS,khan_real-time_2021} or post-processing steps \cite{laroche_fast_2021}.
Today, there is an increasing interest in model-based approaches \cite{IPB_Ozkan,RED_USIPB} that better formalize the problem within an optimization framework. 
%
A second branch of methods for improving US image quality leverages the power of Deep Neural Networks (DNNs). Initial approaches in this direction have been trained to predict B-mode images directly \cite{Hyun19}, the
beamforming weights \cite{luijten_adaptive_2020,MNV2} or used as post-processing denoisers under supervised training schemes~\cite{perdios_cnn-based_2022,AUGAN_2021}. Despite their effectiveness, these methods require datasets of corresponding low-high quality image pairs and therefore do not generalize to other organs/tasks. 

Recent hybrid approaches have focused on improving generalizability by combining the best of the model-based and learning worlds. For instance, 
Chennakeshava et al.\ \cite{Chennakeshava:ius2020} propose an unfolding plane-wave compounding method, while Youn et al. \cite{youn:ius2020} combine deep beamforming with an unfolded algorithm for ultrasound localization microscopy. Our work falls within this hybrid model-based deep learning family of approaches \cite{van_sloun_deep_2020}.

We propose the use of DNN image generators to determine and explore the available solution space for the US image reconstruction problem. In practice, we leverage the recent success of Denoising Diffusion Probabilistic Models (DDPMs) \cite{ho_denoising_2020,nichol_improved_2021,dhariwal_diffusion_2021},  which are the state-of-the-art in image synthesis in the domain of natural images. More specifically, we build on the Denoising Diffusion Restoration Models (DDRMs) framework proposed by Kawar et al.~\cite{kawar_denoising_2022}, which adapts DDPMs to various image restoration tasks modeled as linear inverse problems. The main advantage of DDRMs is exploiting the direct problem modeling to bypass the need to retrain DDPMs when addressing new tasks. While the combination of model-based and diffusion models has been explored in the context of CT/MRI imaging \cite{song_solving_2022}, this is, to the best of our knowledge, the first probabilistic diffusion model approach for ultrasound image reconstruction. 

Our methodological contributions are twofold. First, we adapt DDRMs from restoration tasks in the context of natural images (e.g.\ denoising, inpainting, superresolution), to the reconstruction of B-mode US images from raw radiofrequency RF channel data. Our approach can be applied to different acquisition types, e.g. sequential imaging, synthetic aperture, and plane-wave, as long as the acquisition can be approximately modeled as a linear inverse problem, i.e.\ with a model matrix depending only on the geometry and pulse-echo response (point spread function).  Our second contribution is introducing a whitening step to cope with the direct US imaging model breaking the \textit{i.i.d.} noise assumption implicit in diffusion models. In addition to the theoretical advances, we provide a qualitative and quantitative evaluation of the proposed approach on synthetic data under different noise levels, showing the feasibility of our approach. Finally, we also demonstrate results on the PICMUS dataset. 
Next, we review DDRM and introduce our method in Section~\ref{sec:IPB}.

\section{Denoising Diffusion Restoration Models}
\label{sec:DDRM}
\begin{figure}[h]
\centering
\begin{subfigure}{0.47\textwidth}
    \includegraphics[width=\textwidth]{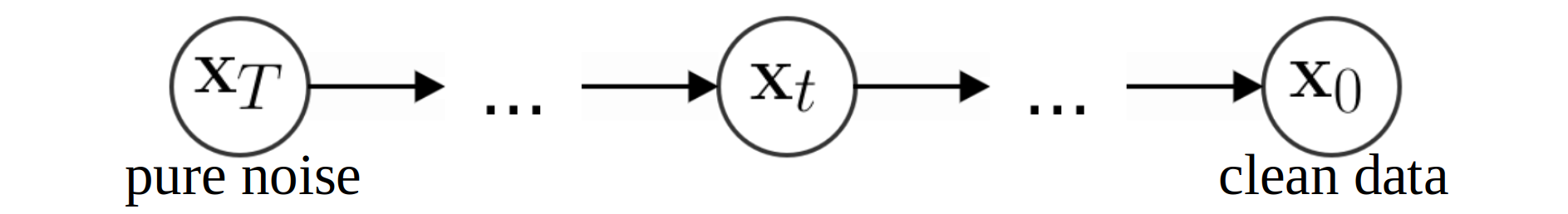}
    \caption{}
    \label{fig: sub_DDPM}
\end{subfigure}
\begin{subfigure}{0.47\textwidth}
    \includegraphics[width=\textwidth]{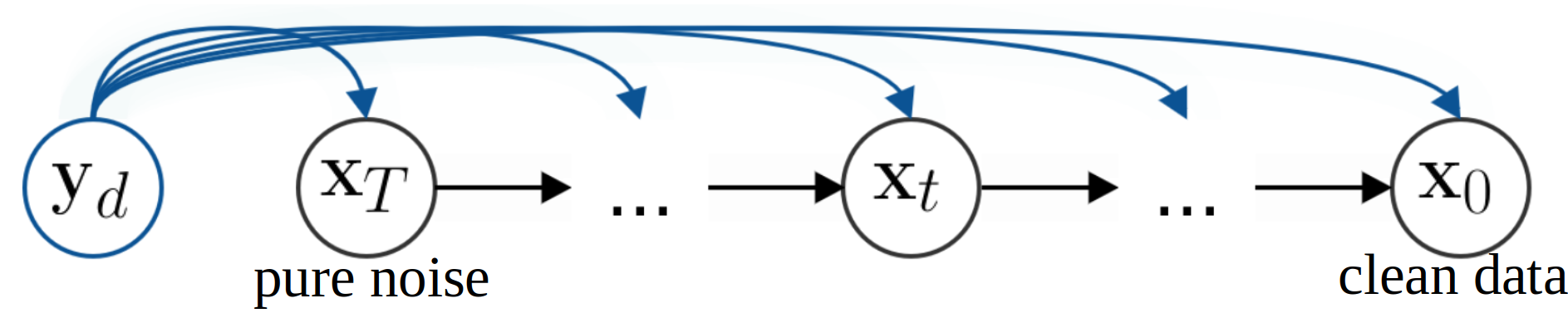}
    \caption{}
    \label{fig: sub_DDRM}
\end{subfigure}        
\caption{
(a) DDPMs vs. (b) DDRMs Generation.
While DDPMs are unconditional generators, DDRMs condition the generated images to measurements $\yv_d$.}
\label{fig: DDRM}
\end{figure}

 \DM{
A DDPM is a parameterized Markov chain trained to generate \ji{synthetic} images from noise relying on variational inference~\cite{ho_denoising_2020,nichol_improved_2021,dhariwal_diffusion_2021}. 
The Markov chain consists }\YZ{of two processes: a forward fixed diffusion process and a backward learned generation process. The forward diffusion process gradually adds Gaussian noise with variance $\sigma_t^2$ ($t = 1,\ldots, T$) to the clean signal $\xv_0$ until it becomes random noise, while in the backward generation process (see Fig.\ref{fig: sub_DDPM}), the random noise $\xv_T$ undergoes a gradual denoising process until a clean $\xv_0$ is generated}.

\DM{An interesting question in model-based deep learning is how}
\YZ{ to use prior knowledge learned by generative models to solve inverse problems.} 
Denoising Diffusion Restoration Models (DDRM)~\cite{kawar_denoising_2022} were recently introduced for solving linear inverse problems, taking advantage of a pre-trained DDPM model as the learned prior. 
\DM{Similar to a DDPM, a DDRM is also a Markov Chain but conditioned on measurements $\yv_d$ through a linear observation model $\Hv_d$  
\footnote{We use subscript $d$ to refer to the original equations of the DDRM model.}. The linear model serves as a link between an unconditioned image generator and any restoration task. In this way, DDRM makes it possible to exploit pre-trained DDPM models \yz{whose weights are assumed to generalize over tasks}. 
\dm{In this sense, DDRM is fundamentally different from previous task-specific learning paradigms requiring training with paired datasets.}
Relying on this principle, the original DDRM paper was shown to work on several}
natural image restoration tasks such as denoising, inpainting, and colorization.

\DM{Different from DDPMs, the Markov chain in DDRM is defined in the spectral space of the degradation operator $\Hv_d$. To this end,}
DDRM leverages the Singular Value Decomposition (SVD): 
$\Hv_d=\Uv_d \Sb_d \Vv_d^\tD$ with $\Sb_d=\Diag\left(s_1,\ldots,s_N\right)$,
\DM{which allows decoupling} the dependencies between the measurements. 
\DM{The original observation model 
$
 \yv_d= \Hv_d \xv_d + \nv_d= \Uv_d \Sb_d \Vv^\tD _d\xv_d + \nv_d,
$
can thus be cast as a denoising problem that can be addressed \yz{on the transformed measurements:}}
\begin{equation*}
\overline{\yv}_d= \overline{\xv}_d + \overline{\nv}_d
\end{equation*}
with $\overline{\yv}_d= \Sb_d^\dag\Uv_d^\tD\yv_d$, $ \overline{\xv}_d=\Vv_d^\tD\xv_d$, and $\overline{\nv}_d=\Sb_d^\dag\Uv_d^\tD \nv_d$, where $\Sb_d^\dag$ is the generalized inverse of $\Sb_d$. 
The additive noise $\nv_d$ being assumed \textit{i.i.d.} Gaussian: $\nv_d\sim \mathcal{N}\left(0, \sigma_d^2\Iv_N\right)$, with a known variance $\sigma_d^2$ and $\Iv_N$ the $N\times N$ identity matrix, we then have
$\overline{\nv}_d$ \yz{with} standard deviation $\sigma_d\Sb_d^\dag$.

Each denoising step from $\overline{\xv}_t$ to $\overline{\xv}_{t-1}$ ($t=T,...,1$) is a linear combination of $\overline{\xv}_t$, the transformed measurements $\overline{\yv}_d$, the transformed prediction of 
$\xv_0$ at the current step $\overline{\xv}_{\theta,t}$, and random noise. To determine their coefficients which are denoted as $A$, $B$, $C$, and $D$ respectively, the condition on the noise, $(A\sigma_t)^2 + (B\sigma_d/s_i)^2 + D^2 = {\sigma_{t-1}}^2$, and on the signal, $A+B+C = 1$, are leveraged, and the two degrees of freedom are taken care of by two hyperparameters.

\yz{In this way, the iterative restoration is achieved by the iterative denoising, and the final restored image is $\xv_0 = \Vv_d\overline{\xv}_{0}$. 
For speeding up this process, skip-sampling \ji{\cite{DDIM}} is applied in practice. We denote the number of iterations as \texttt{it}.}

\section{Method: Reconstructing US images with DDRM}
\label{sec:IPB}
We target the problem of reconstructing US images from raw data towards improving image quality. 
\dm{To model the reconstruction with a linear model,}
we consider the ultrasonic transmission-reception process under the first-order Born approximation.
\ji{We introduce the following notations: $\tau$, $k$, $x$, and $\rv$ respectively denote
the time delay, the time index, the reflectivity function, and the observation position in the field of view.}
When the ultrasonic wave transmitted by the $i^{th}$ element passes through the scattering medium $\Omega$ and is received by the $j^{th}$ element, the received echo signal can be expressed as
\begin{equation}
    y_{i, j}(k) = \int_{\rv \in \Omega}a_i(\rv)a_j(\rv)h(k-\tau_{i,j}(\rv)) x(\rv) \mathrm{d} \rv + n_{j}(k),
\label{Equ: model_continuous}
\end{equation}
where $n_{j}(k)$ represents the noise for the $j^{th}$ receive element, function $h$ \yz{is the convolution of the emitted excitation pulse} and the two-way transducer impulse response, and $a$ represent\dm{s} the weights for apodization according to the transducer's limited directivity.

The discretized linear physical model with $N$ observation points and $K$ time samples for all $L$ receivers can then be 
\dm{rewritten}
as $\yv=\Hv\xv + \nv$, \yz{where $\xv\in \R^{N\times 1}$, $\nv\in \R^{KL\times 1}$, and $\Hv\in \R^{KL\times N}$ is filled with the convolving and multiplying factors from $h$ and $a$ at the delays $\tau_{i,j}$}. 
Due to the \yz{Born approximation, the inaccuracy of $h$ and $a$,} and the discretization, the additive noise $\nv$ \yz{does} not only include the white Gaussian electronic noise but also the model error. However, for simplicity, we still assume $\nv$ as white Gaussian with standard deviation $\gamma$, which is reasonable for the plane wave transmission~\cite{iMAP}. 

While iterative methods exist for solving such \yz{linear inverse} problems~\cite{IPB_Ozkan,RED_USIPB}, our goal is to improve the quality of the reconstructed image by relying on recent advances in diffusion models and, notably, on DDRM.
Given the above linear model, we \yz{can now} rely on DDRM to iteratively guide the reconstruction of the US image from the measurements. However, since DDRM relies on \yz{the} SVD of $\Hv$ to go from a generic inverse problem \yz{to} a denoising/inpainting problem, and \yz{since this} SVD produces huge orthogonal matrices that cannot be implemented as operators, we \yz{propose to} transform the linear inverse problem model to:
\begin{equation}
    \Bv\yv=\Bv\Hv\xv + \Bv\nv,
    \label{Equ: model_HtH}
\end{equation}
where $\Bv \in \R^{N \times KL}$ is a beamforming matrix that \yz{projects channel data to the image domain}. 
\yz{After this transformation, we then feed the new inverse problem (Eq.~\ref{Equ: model_HtH}) to DDRM to iteratively reconstruct $\xv$ from $\Bv\yv$ observations. In this way, the size of the SVD of $\Bv\Hv$ becomes more tractable.}
We call this first model DRUS for Diffusion Reconstruction in US.

However, the noise of the updated direct model $\Bv\nv$ is no longer white and thus, it does not meet the assumption of DDRM. For this reason, we introduce a whitening operator $\Cv \in \R^{M \times N}$, where $M \leqslant N$, and upgrade the inversion model to its final form:
\begin{equation}
    \Cv\Bv\yv = \Cv\Bv\Hv\xv + \Cv\Bv\nv,
    \label{Equ: model_CHtH}
\end{equation}
where \Cv is such that $\Cv\Bv\nv$ is a white noise sequence. In order to compute $\Cv$, we rely on the eigenvalue decomposition 
$
    \Bv\Bv^\tD = \Vv \Lambdab \Vv^\tD
$
where $\Lambdab \in \R ^{N \times N}$ is a diagonal matrix of the eigenvalues of $\Bv\Bv^\tD$, and $\Vv \in \R^{N \times N}$ is a matrix whose columns are the corresponding right eigenvectors. Then, the covariance matrix of the whitened additive noise $\Cv\Bv\nv$ can be written as
\begin{align*}
 \mathrm{Cov}(\Cv\Bv\nv)&= \ED[\Cv\Bv\nv \nv^\tD \Bv^\tD \Cv^\tD]
 = \gamma^2 \Cv\Bv\Bv^\tD\Cv^\tD
 = \gamma^2 \Cv\Vv \Lambdab \Vv^\tD \Cv^\tD.
\end{align*}
Now, let $\Cv =\Pv\Lambdab^{-\frac{1}{2}} \Vv^\tD$ with $\Pv=[\Iv_M,\zerob_{M\times(N-M+1)}] \in \R^{M \times N}$. It can be easily checked that $\Cv\Vv \Lambdab \Vv^\tD \Cv^\tD=\Iv_M$, \yz{proving the noise $\Cv\Bv\nv$ is white}.

\yz{Besides, discarding the smallest eigenvalues by empirically choosing $M$, rather than strictly limiting ourselves to zero eigenvalues, can compress the size of the observation vector $\Cv\Bv\yv$ from $N \times 1$ to $M \times 1$ and make the size of the SVD of $\Cv\Bv\Hv$ more  tractable.} 

In order to adapt DDRM to the final inverse model in Eq.~\ref{Equ: model_CHtH}, we consider $\yv_d=\Cv\Bv\yv$ and $\nv_d=\Cv\Bv\nv$ as input and \yz{compute the SVD of $\Hv_d=\Cv\Bv\Hv$.} We name this whitened version of the approach WDRUS. In summary:
\begin{itemize}
    \item \DM{DRUS model relies on} ~\eqref{Equ: model_HtH} \DM{with} $\yv_d=\Bv\yv$,  $\Hv_d=\Bv\Hv$ and $\nv_d=\Bv\nv$
    \item \DM{WDRUS model relies on}~\eqref{Equ: model_CHtH} \DM{with} $\yv_d=\Cv\Bv\yv$, $\Hv_d=\Cv\Bv\Hv$ and $\nv_d=\Cv\Bv\nv$.
\end{itemize}

\section{Experimental Validation}
\label{sec:results}
In our study, we employed an open-source generative diffusion model~\cite{dhariwal_diffusion_2021} at resolution $256 \times 256$ pre-trained on ImageNet~\cite{ILSVRC15}. We evaluated our method \yz{with $\texttt{it}=50$ on both} synthetic data and on the Plane Wave Imaging Challenge in Medical UltraSound (PICMUS)~\cite{PICMUS} dataset. \yz{For the latter, we also experimented with \yz{the same} unconditional diffusion model \yz{but this time} fine-tuned with 800 high-quality \dm{unpaired} ultrasound images acquired with \yz{a} TPAC Pioneer machine on \yz{a CIRS 040GSE phantom}. Image samples and data acquisition parameters for fine-tuning are in the supplementary material}.

\yz{All evaluations in this paper are performed in plane-wave modality. The baseline for synthetic data is 
beamformed by applying matched filtering $\mathbf{B}=\mathbf{H}^t$. 
\dm{The} references for the PICMUS dataset apply DAS 
with 1, 11, and 75 transmissions. Our proposed DRUS and WDRUS are compared 
\dm{against the DAS references}
by taking \dm{measurements of} a single-transmission 
as input.}

\subsection{Results on synthetic data}\label{sec: synthetic}
We simulate data from two phantoms, \yz{a synthetic \texttt{SynVitro} and Field\,II \texttt{fetus} \cite{FieII_1,FieII_2}}. \CH{The model matrix $\Hv$ includes receive apodization using \CH{Hann} window and \texttt{f-number}\,$=0.5$, and the beamformer $\mathbf{B}=\mathbf{H}^t$.} We simulated channel data $\mathbf{y} = \mathbf{Hx}+\mathbf{n}$ with six levels of additive noise ($\gamma = 0.3, 0.7, 1.0, 1.5, 2.0, 2.5$).

\begin{figure}[h]
\centering
\begin{subfigure}{0.45\textwidth}
    \includegraphics[width=\textwidth]{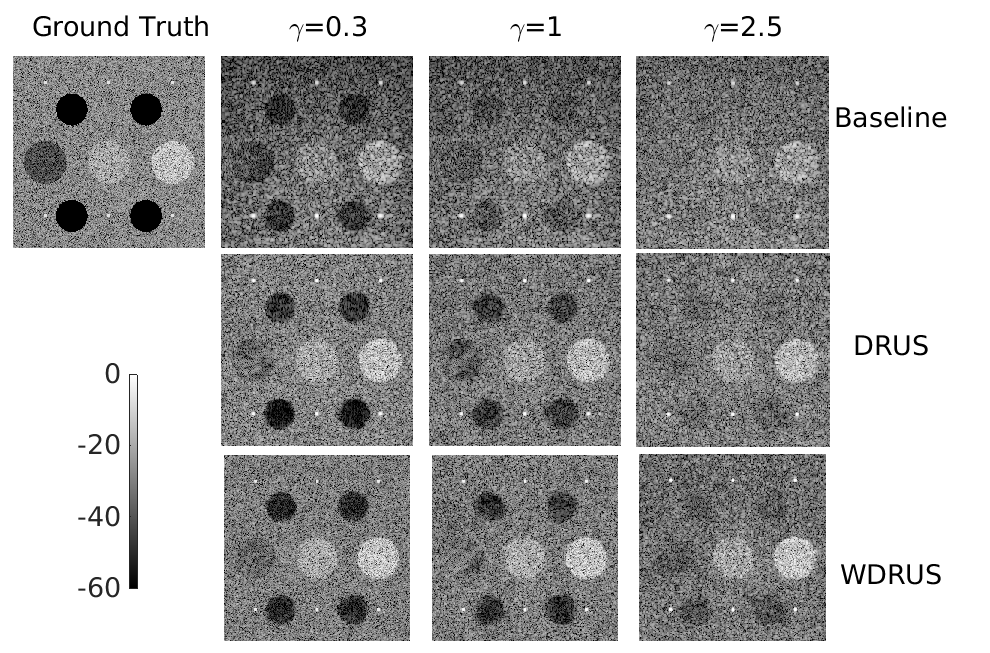}
    \caption{ \yz{\texttt{SynVitro}} phantom.}
    \label{fig: SynVitro_imgs}
\end{subfigure}
\quad
\begin{subfigure}{0.45\textwidth}
    \includegraphics[width=\textwidth]{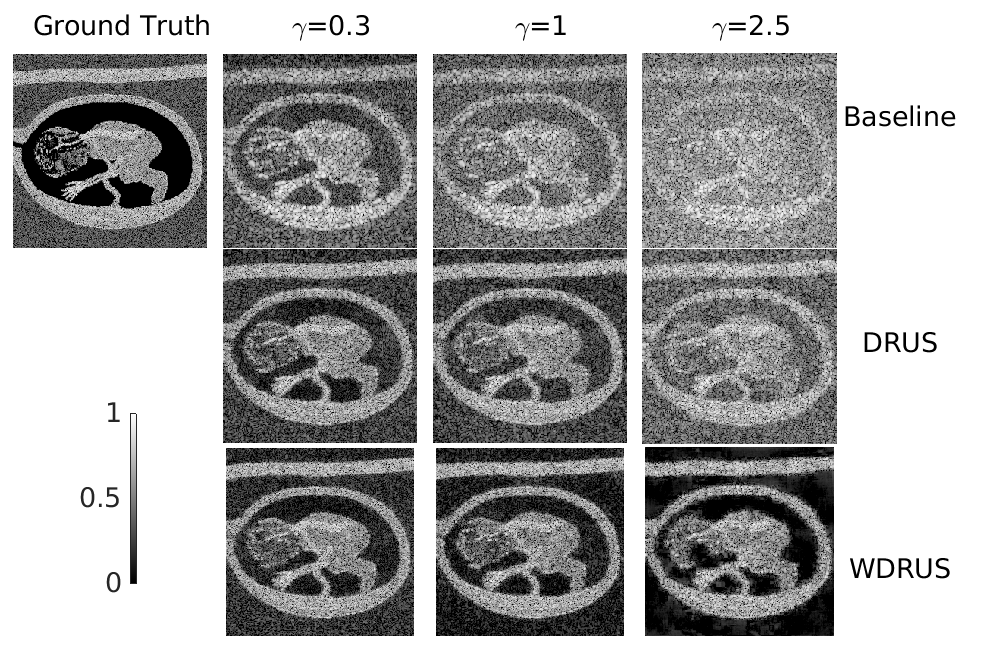}
    \caption{ \yz{\texttt{fetus}} phantom.}
    \label{fig: fetus_imgs}
\end{subfigure}

\begin{subfigure}{\textwidth}
    \includegraphics[width=\textwidth]{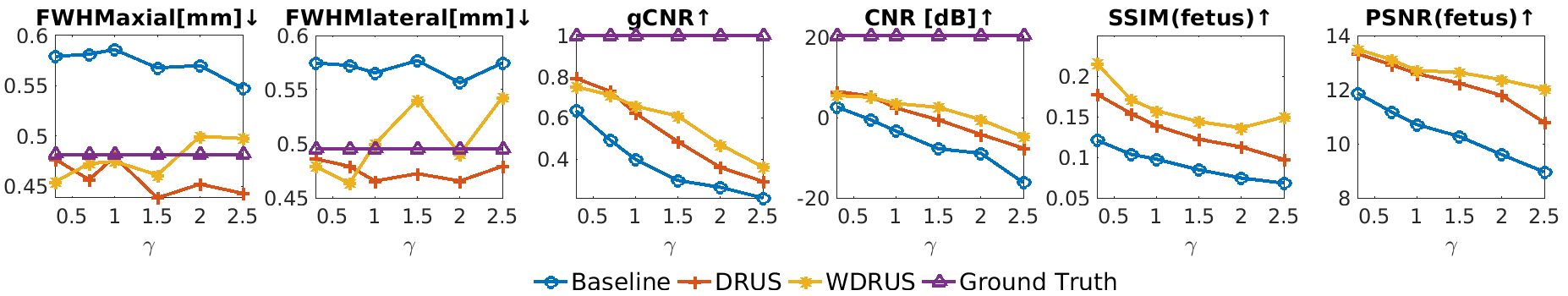}
    \caption{Comparative evaluation of image quality metrics on synthetic data.}
    \label{fig: metrics_synthetic}
\end{subfigure}
\caption{Comparison of restored images on synthetic data.
$\texttt{it}=50$ for DRUS and WDRUS. \yz{All images are in decibels with a dynamic range [-60,0]. The \texttt{fetus} images are normalized between 0 to 1 for calculating the SSIM and PSNR.}}
\label{fig: results_synthetic}
\end{figure}

The restoration quality for \texttt{SynVitro} is quantitatively evaluated with both resolution and contrast metrics. For the \texttt{fetus} phantom, we measure the Structural SIMilarity (SSIM)~\cite{SSIM} and the Peak Signal-to-Noise Ratio (PSNR).
Resolution is measured as the -6dB Full Width at Half Maximum (FWHM) in axial and lateral directions separately on the six bright scatterers. For evaluating the contrast, we rely on both the Contrast to Noise Ratio (CNR) and the generalized Contrast to Noise Ratio (gCNR):
\ji{$$\mathrm{CNR}=10 \log_{10}\bigg(\frac
{\left|\mu_{\text{in}}-\mu_{\text{out}}\right|^2}{\left(\sigma_{\text {in }}^2+\sigma_{\text {out}}^2\right) / 2}\bigg),
\quad
\mathrm{gCNR}=1-\int_{-\infty}^{\infty} \min \left\{f_{\text {in}}(v), f_{\text {out}}(v)\right\} dv,
$$}%
\yz{both measured on the four anechoic regions, where the subscripts `in' and `out' indicate inside or outside the target regions, $v$ denotes the pixel values, and $f$ refers to the histograms of pixels in each region. The restored images and metrics are summarized in Fig.~\ref{fig: results_synthetic}. The metrics for \texttt{SynVitro} are averaged over the different noise levels for simplicity.}

\yz{Qualitatively and quantitatively,} both DRUS and WDRUS significantly outperform the matched-filtering baseline $\Hv^t\yv$, and WDRUS is generally superior to DRUS in terms of noise reduction and contrast enhancement. The two proposed approaches even outperform the ground truth for resolution at low-noise conditions (e.g. $\gamma = 0.3, 0.7, 1.0$), while the resolution of images restored by WDRUS under high-noise conditions (e.g. $\gamma = 1.5, 2.0, 2.5$) is worse than that of DRUS.

\subsection{Results on PICMUS dataset}\label{sec: picmus}
There are four phantoms in the PICMUS~\cite{PICMUS} dataset. 
\yz{\texttt{SR} and \texttt{SC} are Field\,II \cite{FieII_1,FieII_2} simulations while \texttt{ER} and \texttt{EC} were acquired on a CIRS 040GSE phantom.}
\CH{
We use the PICMUS presets where the beamformer $\Bv$ comprises receive apodization using Tuckey25 window and \texttt{f-number}\,$=1.4$, while $\Hv$ has no apodization.
}

In addition to using FWHM, CNR, and gCNR for evaluating resolution (for \texttt{SR} and \texttt{ER}) and contrast (for \texttt{SC} and \texttt{EC}) introduced in Section~\ref{sec: synthetic}, we also use the Signal to Noise Ratio (SNR) $\mu_\text{ROI} / \sigma_\text{ROI}$ and the Kolmogorov–Smirnov (KS) test at the 5$\%$ significance level, for evaluating the speckle quality preservation (for \texttt{SC} and \texttt{EC}), where ROI is the region of interest. SNR $\approx 1.91$ and passing the KS test under a Rayleigh distribution hypothesis are \yz{indicators of a} good speckle texture preservation. 

\begin{figure}[ht]
\centering
\includegraphics[width=0.85\textwidth]{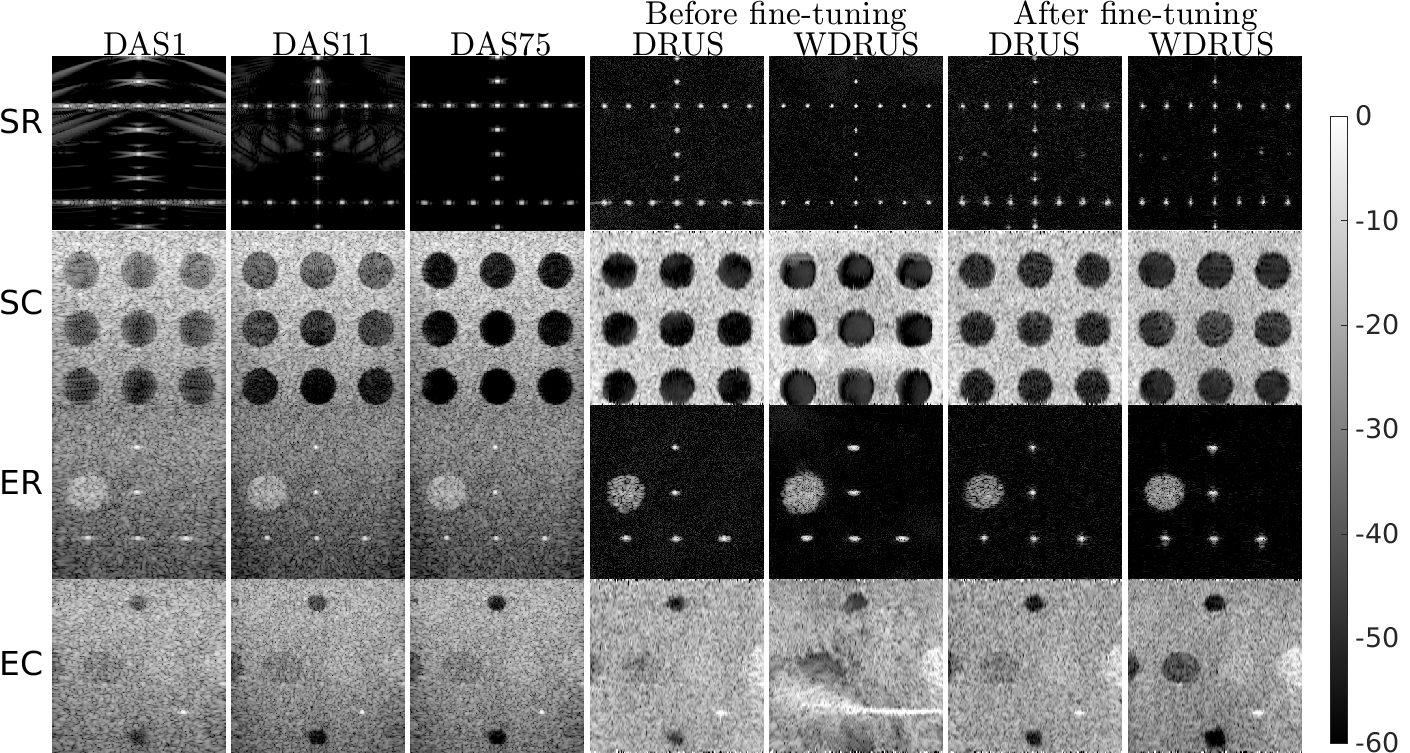}
\caption{Reconstructed images comparison on the PICMUS~\cite{PICMUS} dataset using various approaches. \yz{All images are in decibels with a dynamic range [-60,0].}
} \label{fig: picmus images}
\end{figure}

Using single plane-wave transmission (1PW), we compare our approaches with DAS (1PW, 11PWs, and 75 PWs) qualitatively and quantitatively in Fig. \ref{fig: picmus images} and in Table \ref{Tab: picmus metrics}, respectively. We also compare with the scores (\yz{taken} from~\cite{RED_USIPB}) of four other approaches in Table \ref{Tab: picmus metrics}, including the Eigenspace-based Minimum Variance (EMV)~\cite{asl_eigenspace-based_2010} which does adaptive beamforming, the traditional Phase Coherence Imaging (PCF)~\cite{PCF}, a model-based approach 
with
regularization by denoising (RED)~\cite{RED_USIPB}, and \yz{a learning-based approach} using MobileNetV2 (MNV2)~\cite{MNV2}.

\begin{table}[ht]
\caption{\dm{Image quality metrics on the PICMUS SR, SC, ER, EC datasets.}}
\label{Tab: picmus metrics}
\resizebox{\textwidth}{!}{%
\begin{tabular}{ccc|ccc|cc|cc|cccc}
\hline
 & \multicolumn{2}{|c|}{Metric}  & \multicolumn{3}{c|}{DAS} & \multicolumn{2}{c|}{no fine-tuning} & \multicolumn{2}{c|}{after fine-tuning} & \multirow{2}{*}{EMV\cite{asl_eigenspace-based_2010}} & \multirow{2}{*}{PCF\cite{PCF}} & \multirow{2}{*}{RED\cite{RED_USIPB}} & \multirow{2}{*}{MNV2\cite{MNV2}} \\ \cline{7-10}
 & \multicolumn{2}{|c|}{}  & 1 & 11 & 75 & DRUS & WDRUS & DRUS & WDRUS &  &  &  &  \\ \hline
\multicolumn{1}{c|}{\multirow{2}{*}{SR}} & \multirow{2}{*}{\begin{tabular}[c]{@{}c@{}}FWHM\\ {[}mm{]}\end{tabular}} & A$\downarrow$ & 0.38 & 0.38 & 0.38 & 0.30 & 0.32 & 0.34 & 0.31 & 0.40 & 0.30 & 0.37 & 0.42 \\
\multicolumn{1}{c|}{} &  & L$\downarrow$ & 0.81 & 0.53 & 0.56 & 0.47 & 0.31 & 0.39 & 0.28 & 0.10 & 0.38 & 0.46 & 0.27 \\ \hline
\multicolumn{1}{c|}{\multirow{4}{*}{SC}} & \multicolumn{2}{c|}{CNR{[}dB{]}$\uparrow$} & 10.41 & 12.86 & 15.89 & 16.37 & 15.20 & 15.74 & 16.33 & 11.21 & 0.46 & 15.48 & 10.48 \\
\multicolumn{1}{c|}{} & \multicolumn{2}{c|}{gCNR$\uparrow$} & 0.91 & 0.97 & 1.00 & 0.99 & 0.99 & 0.99 & 0.99 & 0.93 & 0.41 & 0.94 & 0.89 \\
\multicolumn{1}{c|}{} & \multicolumn{2}{c|}{SNR $|$ KS} & 1.72$|$\Checkmark & 1.69$|$\Checkmark & 1.68$|$\Checkmark & 2.06$|$\Checkmark & 1.98$|$\Checkmark & 2.03$|$\Checkmark & 1.99$|$\Checkmark & / $|$ \Checkmark & / $|$ \XSolidBrush & / $|$ \Checkmark & / $|$ \Checkmark \\ \hline
\multicolumn{1}{c|}{\multirow{2}{*}{ER}} & \multirow{2}{*}{\begin{tabular}[c]{@{}c@{}}FWHM\\ {[}mm{]}\end{tabular}} & A$\downarrow$ & 0.56 & 0.54 & 0.54 & 0.34 & 0.34 & 0.27 & 0.22 & 0.59 & 5.64 & 0.48 & 0.53 \\
\multicolumn{1}{c|}{} &  & L$\downarrow$ & 0.87 & 0.54 & 0.56 & 0.63 & 1.05 & 0.55 & 0.69 & 0.42 & 0.76 & 0.76 & 0.77 \\ \hline
\multicolumn{1}{c|}{\multirow{4}{*}{EC}} & \multicolumn{2}{c|}{CNR{[}dB{]}$\uparrow$} & 7.85 & 11.20 & 12.00 & 9.00 & -7.25 & 11.75 & 13.55 & 8.10 & 3.20 & 14.70 & 7.80 \\
\multicolumn{1}{c|}{} & \multicolumn{2}{c|}{gCNR$\uparrow$} & 0.87 & 0.94 & 0.95 & 0.88 & 0.69 & 0.96 & 0.97 & 0.83 & 0.68 & 0.98 & 0.83 \\
\multicolumn{1}{c|}{} & \multicolumn{2}{c|}{SNR $|$ KS} & 1.97$|$\Checkmark & 1.91$|$\Checkmark & 1.92$|$\Checkmark & 1.91$|$\Checkmark & 1.50$|$\XSolidBrush & 2.11$|$\Checkmark & 1.92$|$\Checkmark & / $|$ \Checkmark & / $|$ \XSolidBrush & / $|$ \Checkmark & / $|$ \Checkmark \\ \hline
\end{tabular}%
}
\end{table}

In terms of resolution and contrast, our method is overall significantly better than DAS with 1 plane-wave transmission and can compete with DAS with 75 plane-wave transmissions, as seen in Table.~\ref{Tab: picmus metrics}. However, when the diffusion model is \yz{not fine-tuned (using the pre-trained weights from ImageNet \cite{dhariwal_diffusion_2021}), 
artifacts} on the \texttt{EC} image \dm{are} recovered by WDRUS, which can be explained from two perspectives.
\yz{First, while a pre-trained model is a powerful prior and frees the user from acquiring data and training a huge model, there is still a gap between the distribution of natural vs. ultrasound images.} This point can be confirmed by comparing the performance of DRUS and WDRUS in Fig.\ref{fig: picmus images} before [col(4,5)] and after [col(6,7)] fine-tuning the diffusion model. With the latter, both DRUS and WDRUS reconstruct images with less distortion, particularly for the anechoic regions on \texttt{SC} and \texttt{EC}, and the hyperechoic region on \texttt{ER}.

Second, due to the approximation of the impulse responses, matrix B has a certain degree of error \yz{which is propagated through the eigenvalue decomposition of B and the whitening matrix C. Eventually, such errors may lead to a larger error in WDRUS than in DRUS, as seen when comparing WDRUS [col(5,7)] and DRUS [col(4,6)], despite better contrast and SSIM metrics in Fig.~\ref{fig: metrics_synthetic}. These errors may also explain why WDRUS is weaker than DRUS in terms of lateral resolution (FWHM L in Table.~\ref{Tab: picmus metrics}) of scatterers in the \texttt{ER} phantom.}

\yz{Finally,} although our method can reconstruct high-quality \texttt{SR}, \texttt{SC}, and \texttt{EC} images using the fine-tuned diffusion model, it is still difficult to retain speckle quality for \texttt{ER}, 
\dm{which is a current limitation. }

\section{Discussion and Conclusion}
\label{sec:conclusion}


Regarding the computing time, \yz{our approaches need 3-4 minutes to form one image, which is slower than DAS1, PCF~\cite{PCF} and MNV2~\cite{MNV2}, but faster \ji{than EMV ~\cite{asl_eigenspace-based_2010} and RED~\cite{RED_USIPB}, which need 8 and 20 minutes, respectively.} RED is slow because each iteration contains an inner iteration while \ji{EMV spends time on covariance matrix evaluation and decomposition.} Our iteration restoration approaches require multiple multiplication operations with the singular vector matrix, which currently hinders real-time imaging. }Accelerating this process is one of our key focuses for future work.

\YZ{In conclusion, for the first time, we achieve the reconstruction of ultrasound images with} 
\DM{ two adapted diffusion models, DRUS and WDRUS. }
\yz{Different from previous model-based deep learning methods which are task-specific and require a large amount of data pairs for supervised training, our approach requires none or just a small fine-tuning dataset composed of high-quality (e.g., DAS101) images only (there is no need for paired data). Furthermore, the fine-tuned diffusion model can be used}
\dm{for other US related inverse problems.}
\YZ{Finally, our method demonstrated competitive performance compared to DAS75, and other state-of-the-art approaches on the PICMUS dataset.}

\bibliographystyle{splncs04}
\bibliography{bibliography}
\newpage
\thispagestyle{empty}



\begin{table}[H]
\caption{Parameters for data acquisition and fine-tuning}
\resizebox{\textwidth}{!}{%
\begin{tabular}{@{}lccc@{}}
\toprule
\multicolumn{4}{c}{Parameters} \\ \midrule
\multicolumn{2}{c|}{acquisition} & \multicolumn{2}{c}{training} \\ \midrule
\multicolumn{1}{l|}{transmission modality} & \multicolumn{1}{c|}{Plane Wave} & \multicolumn{1}{c|}{format} & h5 \\
\multicolumn{1}{l|}{number of plane waves} & \multicolumn{1}{c|}{101 per image} & \multicolumn{1}{c|}{ random flip up/down} & True \\
\multicolumn{1}{l|}{angles} & \multicolumn{1}{c|}{[$\alpha_{min}$, $\alpha_{step}$, $\alpha_{max}$] = [-25$^{\circ}$, 0.5$^{\circ}$, 25$^{\circ}$]} & \multicolumn{1}{c|}{random flip left/right} & True \\
\multicolumn{1}{l|}{center frequency} & \multicolumn{1}{c|}{5 MHz} & \multicolumn{1}{c|}{random reverse values} & True \\
\multicolumn{1}{l|}{Digital Gain } & \multicolumn{1}{c|}{60} & \multicolumn{1}{c|}{number of images} & $103 \times 8 = 824$ \\
\multicolumn{1}{l|}{time range} & \multicolumn{1}{c|}{70 {[}$\mu$s{]}} & \multicolumn{1}{c|}{name of the saved data} & CIRS [11-18] \\
\multicolumn{1}{l|}{software TGC} & \multicolumn{1}{c|}{0.008 (quadrature)} & \multicolumn{1}{c|}{} &  \\
\multicolumn{1}{l|}{f-number} & \multicolumn{1}{c|}{1} & \multicolumn{1}{c|}{} &  \\
\multicolumn{1}{l|}{receive apodization} & \multicolumn{1}{c|}{tukey 25} & \multicolumn{1}{c|}{} &  \\
\multicolumn{1}{l|}{sampling frequency} & \multicolumn{1}{c|}{50 MHz} & \multicolumn{1}{c|}{} &  \\ \bottomrule
\end{tabular}%
}
\label{tab: dataAcquisition_params}
\end{table}

\begin{figure}[H]
    \centering
    \includegraphics[width=\textwidth]{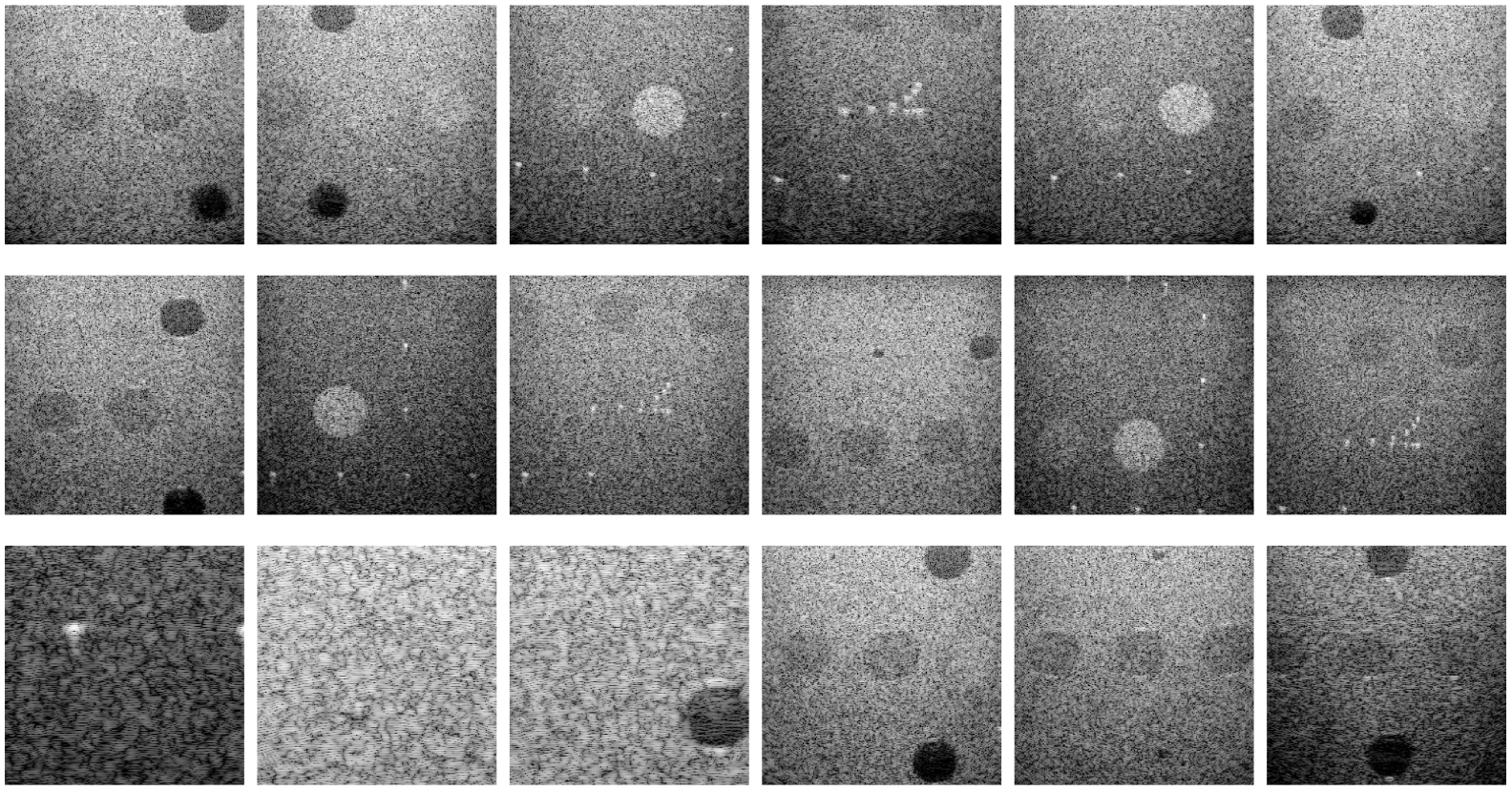}
    \caption{Examples of the fine-tune set}
    \label{fig: fine-tune samples}
\end{figure}
\end{document}